\documentclass[conference]{IEEEtran}
\usepackage{xcolor}
\usepackage{blindtext}
\usepackage{etoolbox}
\makeatletter
\patchcmd{\@makecaption}
  {\scshape}
  {}
  {}
  {}
\makeatletter
\patchcmd{\@makecaption}
  {\\}
  {.\ }
  {}
  {}
\makeatother

\usepackage{siunitx}

\usepackage{cite}
\usepackage[pdftex]{graphicx}
  \graphicspath{{../pdf/}{../jpeg/}}
  \DeclareGraphicsExtensions{.pdf,.jpeg,.png}
\usepackage{amsmath}
\usepackage{txfonts}
\DeclareMathOperator*{\argmin}{arg\,min}

\usepackage{array}
\newcolumntype{U}{>{\centering\arraybackslash}m{0.75cm}}
\newcolumntype{V}{>{\centering\arraybackslash}m{4.5cm}}
\ifCLASSOPTIONcompsoc
  \usepackage[caption=false,font=normalsize,labelfont=sf,textfont=sf]{subfig}
\else
  \usepackage[caption=false,font=footnotesize]{subfig}
\fi
\usepackage{url}
\usepackage{makecell}

\begin{document}
%\title{Unsupervised Rotation Invariant Moveme Learning from Static Poses}
\title{A Rotation Invariant Latent Factor Model for Moveme Discovery from Static Poses}

\author{

\IEEEauthorblockN{Matteo Ruggero Ronchi, Joon Sik Kim and Yisong Yue}
\IEEEauthorblockA{California Institute of Technology, Pasadena, CA, USA\\
\texttt{\{mronchi, jkim5, yyue\}@caltech.edu}\\}
\vspace{5mm}
% \and

% \IEEEauthorblockN{Anonymous Author 2}
% \IEEEauthorblockA{Anonymous Institution 2\\
% Anonymous Address 2\\
% Anonymous Email 2}

% \and

% \IEEEauthorblockN{Anonymous Author 3}
% \IEEEauthorblockA{Anonymous Institution 3\\
% Anonymous Address 3\\
% Anonymous Email 3}

}

\maketitle


%%%%%%%%%%%%%%%%%%%%%%%%%%%%%%%%%%%%%%%%%%%%%%%%%%%%%%%%%%%%%%%%%%%%%%%%%%%%%%%%%%%%%%%%%%
%%%%%%%%%%%%%%%%%%%%%%%%%%%%%%%%%%%%%%%%%%%%%%%%%%%%%%%%%%%%%%%%%%%%%%%%%%%%%%%%%%%%%%%%%%
%% ABSTRACT
%%%%%%%%%%%%%%%%%%%%%%%%%%%%%%%%%%%%%%%%%%%%%%%%%%%%%%%%%%%%%%%%%%%%%%%%%%%%%%%%%%%%%%%%%%
%%%%%%%%%%%%%%%%%%%%%%%%%%%%%%%%%%%%%%%%%%%%%%%%%%%%%%%%%%%%%%%%%%%%%%%%%%%%%%%%%%%%%%%%%%

\begin{abstract}
We tackle the problem of learning a rotation invariant latent factor model when the training data is comprised of lower-dimensional projections of the original feature space. The main goal is the discovery of a set of 3-D bases poses that can characterize the manifold of primitive human motions, or movemes, from a training set of 2-D projected poses obtained from still images taken at various camera angles. The proposed technique for basis discovery is data-driven rather than hand-designed. The learned representation is rotation invariant, and can reconstruct any training instance from multiple viewing angles. We apply our method to modeling human poses in sports (via the Leeds Sports Dataset), and demonstrate the effectiveness of the learned bases in a range of applications such as activity classification, inference of dynamics from a single frame, and synthetic representation of movements.
\end{abstract}

%%%%%%%%%%%%%%%%%%%%%%%%%%%%%%%%%%%%%%%%%%%%%%%%%%%%%%%%%%%%%%%%%%%%%%%%%%%%%%%%%%%%%%%%%%
%%%%%%%%%%%%%%%%%%%%%%%%%%%%%%%%%%%%%%%%%%%%%%%%%%%%%%%%%%%%%%%%%%%%%%%%%%%%%%%%%%%%%%%%%%
%% Introduction
%%%%%%%%%%%%%%%%%%%%%%%%%%%%%%%%%%%%%%%%%%%%%%%%%%%%%%%%%%%%%%%%%%%%%%%%%%%%%%%%%%%%%%%%%%
%%%%%%%%%%%%%%%%%%%%%%%%%%%%%%%%%%%%%%%%%%%%%%%%%%%%%%%%%%%%%%%%%%%%%%%%%%%%%%%%%%%%%%%%%%

\section{Introduction}

%% PROBLEM STATEMENT
What are the typical ranges of motion for human arms? What types of leg movements tend to correlate with specific shoulder positions? How can we expect the arms to move given the current body pose? Our goal is to address these questions by recovering a set of ``bases poses'' that summarize the variability of movements in a given collection of static poses captured from images at various viewing angles. 

%% MOVEMES, ACTIONS AND ACTIVITIES
One of the main difficulties of studying human movement is that it is a priori unrestricted, except for physically imposed joint angle limits which have been studied in medical text books, typically for a limited number of configurations \cite{grood1988limits, hatze1997three}. Furthermore, human movement may be distinguished into movemes, actions, and activities \cite{anderson2014toward, bregler1997learning} depending on structure, complexity, and duration. Movemes refer to the simplest meaningful pattern of motion: a short, target-oriented trajectory, that cannot be further decomposed, e.g. ``reach'', ``grasp'', ``step'', ``kick''. A complex gesture should be composed out of simple movemes: we define an action as a predefined and ordered sequence of movemes, such as ``drink from a glass'', or ``open a door''. % (reach the door, grasp its handle, lower the hand, etc.). 
An activity is a (possibly stochastic) combination of actions taking place over a stretch of time with a typical and yet variable structure, e.g. ``dine'', ``read''.  
%% IMPORTANCE OF THIS METHOD
Extensive studies have been carried out on human action and activity recognition \cite{poppe2010survey,turaga2008machine}, however little attention has been paid to movemes since human behaviour is difficult to analyze at such a fine scale of dynamics.\footnote{The extent in time and complexity of human motion is not directly observable in still images but requires videos of humans involved in activities which cannot be recorded extensively without legal or ethical issues, as opposed to fly or mouse behaviour which is very well documented \cite{bio2}.} % \cite{bio1,bio2,bio3}; and secondly because it is heavily subject to interpretation.
In this paper, our primary goal is to learn a basis space to smoothly capture movemes from a collection of two dimensional images, although our learned representation can also aid in higher level reasoning. % such as activity analysis (as shown in our experiments).

\begin{figure}[t!]
\centering
\includegraphics[width=0.8\linewidth]{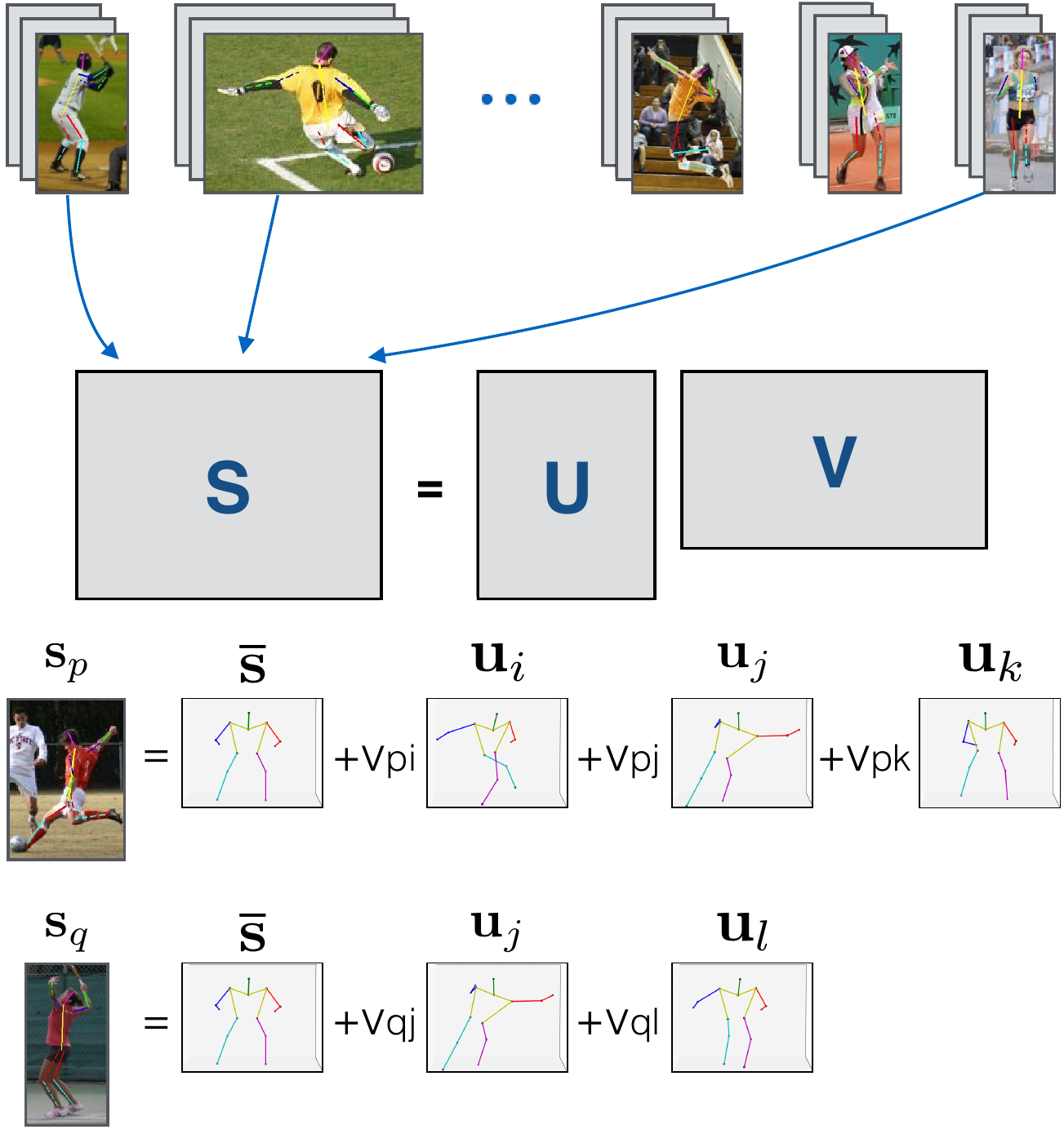}
\caption{ {\small \textbf{Rotation Invariant Moveme Discovery}. Given a collection of static joint locations from images taken at any angle of view we learn a factorization into a basis pose matrix $\mathbf{U}$ and a coefficient matrix $\mathbf{V}$. The learned bases poses in $\mathbf{U}$ are \textit{rotation-invariant} and can be globally applied across a range of viewing angles.
%are global across the whole collection and rotation invariant, as they capture the change in pose underlying a specific movement regardless of the angle of view at which it is portrayed. 
A sparse linear combination of the learned bases accurately reconstructs the pose of a human involved in an action at any angle of view, also for poses not contained in the training set.}}
\label{fig:introduction}
\vspace{-3mm}
\end{figure}

Static poses extracted from two-dimensional images are the most abundant source of pose information. Thus, finding a basis representation using such training data can prove extremely valuable, given the number of image datasets (as opposed to video or mo-cap data) that are currently being collected with a focus on common activities \cite{andriluka14cvpr,BMVC2015_52,krishnavisualgenome}. However, such images are typically taken from a wide range of viewing angles, and can yield only two-dimensional projections of the underlying three-dimensional pose. Any method that does not directly address these issues will learn a naive representation that fails to provide a set of global three-dimensional bases poses that can capture pose changes due to the true human motion while disregarding those due to a change of the angle of view.

In this paper, we propose a simple but effective rotation invariant latent factor model that can recover a set of three-dimensional bases poses from a training set of two-dimensional projections. Our approach is distinguished from previous latent factor modeling approaches by directly incorporating geometric operations in an integrated way, and yields interpretable three dimensional bases poses that can be easily visualized as well as manipulated to express a natural range of human poses (as depicted in Fig.~\ref{fig:introduction}).
We applied our approach in a case study on modeling poses that arise in sports activities, since they have very characteristic and recognizable motions and typically share trajectories of parts of the body (e.g., tennis serve and volleyball strike), which allows to more easily interpret and evaluate qualitatively the learned movemes.

%% APPLICATIONS
Our study is not purely academic, we have four applications in mind; in this paper we carry out a quantitative and qualitative analysis for two of them, and leave the study of the latter to future work. \textbf{Activity recognition}; a compact representation such as the proposed one can be used in addition to the feature representation of state of the art methods for activity recognition, favoring both the performance \cite{yao2011does}, and the interpretability of results. \textbf{Action dynamics inference}; modifying the weights of the learned bases poses is analogous to moving along a line in the high-dimensional space of human poses (either 2-D or 3-D). This allows to predict the future dynamic of an action \cite{fouhey2014predicting}, or morph a pose into another from a single frame, by observing the dynamics of the movemes which better describe the captured pose. \textbf{Computer graphics animation}; many animation systems are still based on \textit{key-framing} and \textit{in-betweening} \cite{parent2012computer}: master animators draw the key frames of a sequence to be animated and assistant animators complete the intermediate frames by inferring the movements occurring between the keys. Knowing the movemes underlying human actions would provide an automated method for interpolating between key frames, resulting in a faster and simplified animation pipeline. \textbf{3-D pose estimation}; a sparse overcomplete dictionary of human poses has been used effectively for the reconstruction of 3-D human pose given its 2-D joint locations from a single frame image \cite{ramakrishna2012reconstructing,fan2014pose,akhter2015pose}. Our technique would allow to identify the most suited pose bases for a given collection of images without any experimenter bias, or the need of curating the angle of view of the images in the training set.\\[-7pt]

%% METHOD CONTRIBUTIONS
In summary, the main contributions of our paper are:

{\bf 1.} An \textbf{unsupervised} method for learning a \textbf{rotation-invariant} set of bases poses. We propose a solution to the intrinsically ill-posed problem of going from static poses to movements, without being affected by the angle of view.

{\bf 2.} A demonstration of how  the learned bases poses can be used in various applications, including manifold traversal, discriminative classification, and synthesis of movements.

%%%%%%%%%%%%%%%%%%%%%%%%%%%%%%%%%%%%%%%%%%%%%%%%%%%%%%%%%%%%%%%%%%%%%%%%%%%%%%%%%%%%%%%%%%
%%%%%%%%%%%%%%%%%%%%%%%%%%%%%%%%%%%%%%%%%%%%%%%%%%%%%%%%%%%%%%%%%%%%%%%%%%%%%%%%%%%%%%%%%%
%% Related Work
%%%%%%%%%%%%%%%%%%%%%%%%%%%%%%%%%%%%%%%%%%%%%%%%%%%%%%%%%%%%%%%%%%%%%%%%%%%%%%%%%%%%%%%%%%
%%%%%%%%%%%%%%%%%%%%%%%%%%%%%%%%%%%%%%%%%%%%%%%%%%%%%%%%%%%%%%%%%%%%%%%%%%%%%%%%%%%%%%%%%%

\section{Related Work}
\subsubsection*{Human Pose Analysis}
There are two main directions of research for human pose analysis. The first one is estimation: given a picture containing a person, the goal is to predict the location of a predefined set of joints of its body, either in the 2-D image \cite{burgos2013robust,chen2014articulated} or in the 3-D space \cite{akhter2015pose,fan2014pose,ramakrishna2012reconstructing}. Methods for 3-D pose reconstruction build upon the results of 2-D pose estimators by using mechanisms based on physical constraints and domain knowledge to infer the true underlying human pose observed in an image, and are more of interest in this study since they implicitly learn an overcomplete basis for modeling human movement. However, such methods typically predefine the dictionary of actions, use additional data in the training phase (such as mo-cap), and do not treat explicitly the problem of varying angles of view. In contrast, our goal is to learn a low-rank manifold of 3-D poses consistent across multiple viewing angles, given only two-dimensional data. %These two approaches are complementary and can be integrated together, as we discuss in the future work.

The second line of investigation uses pose as a form of contextual information that can be combined with objects' category and location in an image to obtain higher performance for activity recognition through a joint learning procedure~\cite{maji2011action,eweiwi2014efficient,yao2010modeling}. Our approach can as well be used as a feature representation for improved activity recognition.%, which we verify empirically.

From the perspective of pose analysis, the goal of this work is to learn a semantically meaningful representation of human pose that can model human motion. This representation should be independent of the application domain, and flexible, allowing it to be incorporated with other representations. Other people investigated this problem: it is known that dynamic information can be recovered from static images of humans engaged in activities \cite{kourtzi2000activation}, and similar representations for action recognition have been learned using video data \cite{kim2010sparse,raptis2013poselet}. We are the first to propose a representation that directly treats the problem of rotation-invariance and can be learned only from static poses, which we believe is important since it is the most abundant form of data.

\subsubsection*{Latent Factor Models and Representation Learning} 
We build upon a long line of research in latent factor models, first popularized for collaborative filtering problems in content recommendation \cite{koren2009matrix}. Applications include modeling variations of faces \cite{turk1991face}, document and text analysis \cite{dumais2004latent}, and behavior patterns in sports \cite{yue14icdm}, amongst many others. Latent factor models are variants of matrix and tensor factorization, which can easily incorporate missing values or other types of constraints. In this regard, our work introduces an approach for learning a latent factor model in a high-dimensional space, when the observed training data are lower-dimensional projections. Our method is complementary to and can be integrated with other latent factor modeling approaches.

Our approach can be viewed as a form of representation learning, which includes methods such as deep neural networks and dictionary learning \cite{bengio2013representation,mairal2009online}.
One of the benefits of representation learning is the ability to smoothly traverse the representation space \cite{gardner2015deep}, which in our setting translates to learning movemes as transitions between poses.

%%%%%%%%%%%%%%%%%%%%%%%%%%%%%%%%%%%%%%%%%%%%%%%%%%%%%%%%%%%%%%%%%%%%%%%%%%%%%%%%%%%%%%%%%%
%%%%%%%%%%%%%%%%%%%%%%%%%%%%%%%%%%%%%%%%%%%%%%%%%%%%%%%%%%%%%%%%%%%%%%%%%%%%%%%%%%%%%%%%%%
%% Methods
%%%%%%%%%%%%%%%%%%%%%%%%%%%%%%%%%%%%%%%%%%%%%%%%%%%%%%%%%%%%%%%%%%%%%%%%%%%%%%%%%%%%%%%%%%
%%%%%%%%%%%%%%%%%%%%%%%%%%%%%%%%%%%%%%%%%%%%%%%%%%%%%%%%%%%%%%%%%%%%%%%%%%%%%%%%%%%%%%%%%%

\section{Models} \label{sec:models}

We develop our approach by building from the classical singular value decomposition. We characterize the challenge of learning only from lower-dimensional projections of the underlying feature space, and present a rotation-invariant latent factor model for dealing with such training data. %., e.g, as used for eigenfaces \cite{turk1991face}.  

\subsection{Basic Notation and Framework} \label{sec:notation}
In this paper, we focus on learning from two-dimensional projections of three-dimensional human poses, however, it is straightforward to generalize to other settings. We are given a training set $S=\{(\mathbf{x}_j,\mathbf{y}_j)\}_{j=1}^{n}$ of $n$ two-dimensional poses, where $x$ and $y$ correspond to the image coordinates of the pose joints from the observed viewing angle, see Fig. \ref{fig:movemes}.  Let $\mathbf{S} \in \Re^{2d \times n}$ denote the dataset matrix, where $2d$ is the dimensionality of the projected space (twice the number of joints $d$ for two-dimensional projections). Our goal is to learn a bases poses matrix $\mathbf{U} \in \Re^{2d \times k}$ composed of $k$ latent factors, and a coefficient matrix $\mathbf{V} \in \Re^{k \times n}$, so that every training example can be represented as a linear combination:
\begin{eqnarray}
\mathbf{s}_j = \mathbf{U}\cdot \mathbf{v_j} + \mathbf{\bar{s}},\label{eqn:basic}
\end{eqnarray}
where $\mathbf{\bar{s}}$ denotes the ``mean'' pose.  %The underlying idea is that the complex movements contained in the dateset can be approximated locally by a linear combination of our the learned movemes.  
Of course, \eqref{eqn:basic} does not deal with rotation invariance and treats the $x$ and $y$ coordinates as having the same semantics across training examples. We present in Sec.~\ref{sec:lfa} a rotation-invariant  latent factor model to address this issue and recover a three-dimensional $\mathbf{U} \in \Re^{3d \times k}$.

%implement rotation invariance in the 2-D case we represent matrices at a discretized angle of view, with $p$ partitions. We denote the subset of the columns of a matrix that belong in a partition with parenthesis: \eg, $\mathbf{S}(a) = [\mathbf{S}_{-,j}]$, $j \in \theta(a)$, where $\theta(a)$ is is the set of indices corresponding to examples in the training set that have angle of view contained in partition $a$.

\subsection{Baselines} \label{sec:baselines}
To the best of our knowledge, no existing approach tackles the problem of learning a rotation-invariant bases for modeling human movement. Previous work is focused on either learning bases poses only from frontal viewing angles or by extensive manual crafting of a predefined set of poses \cite{akhter2015pose,ramakrishna2012reconstructing}. As such, we develop our approach by building upon classical baselines such as the SVD, which we briefly describe here.\\

\subsubsection*{Singular Value Decomposition} The  example in \eqref{eqn:basic} is the most basic form of a latent factor model. When the training objective is to minimize the squared reconstruction error of the training data, then the solution can be recovered via SVD, also used for eigenfaces \cite{turk1991face}.
The bases matrix $\mathbf{U}$ and the coefficient matrix $\mathbf{V}$ respectively correspond (up to a scaling) to the left and right singular vectors of the mean-centered data matrix $\mathbf{S}_{c} = (\mathbf{S} - \mathbf{\bar{s}})$.  
%encoding the amount of displacement of each pose from the mean pose of the dataset, and are obtained by the eigenvector decomposition of $\mathbf{S}_{c}\mathbf{S}_{c}^{T}$ and $\mathbf{S}_{c}^{T}\mathbf{S}_{c}$. This returns the set of orthogonal vectors which best describes the distribution of the data. 
However, naively applying the SVD to our setting will result in the bases matrix $\mathbf{U}$ conflating viewing angle rotations with true pose deformations.\\

\begin{figure}[!t]
\centering
\includegraphics[width=\linewidth]{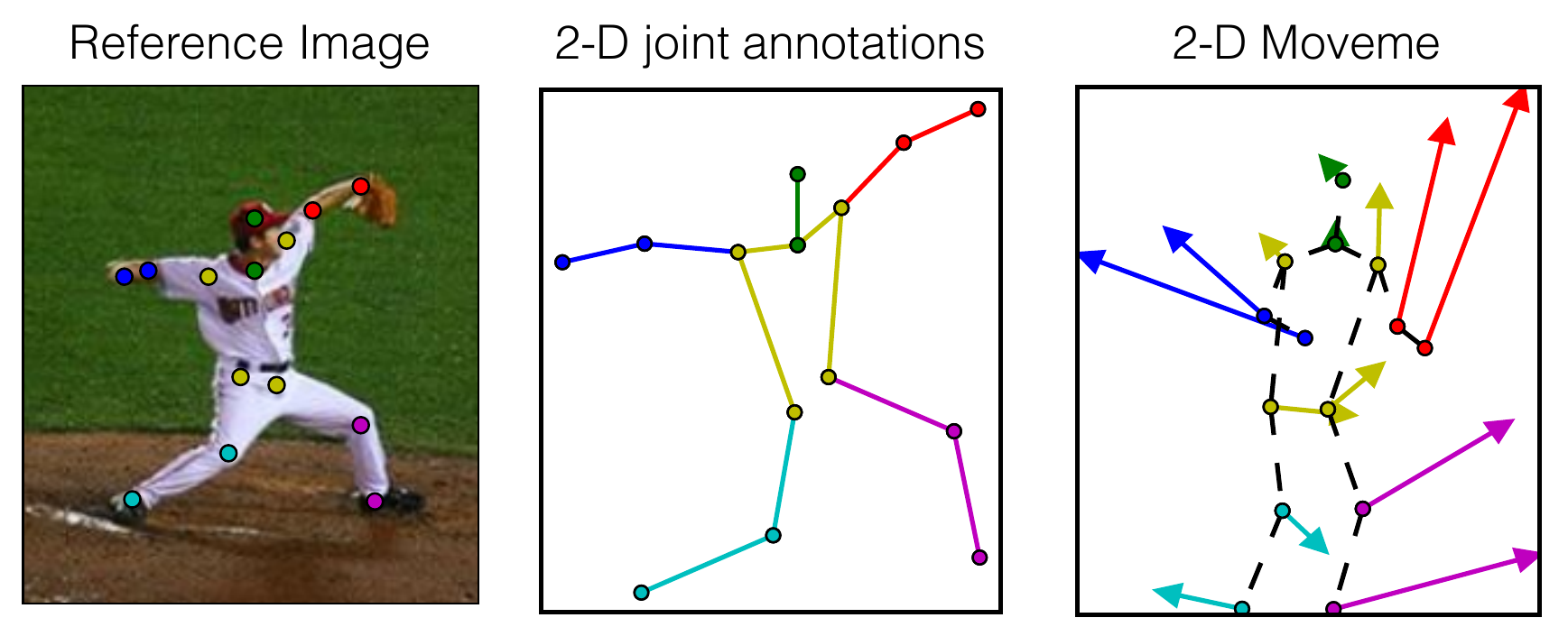}
\caption{ {\small \textbf{Moveme Representation}. The joint annotations from an image in LSP \cite{Johnson10}, and their displacement from the mean pose, which we use to encode movemes.}} 
%A baseball pitch shares trajectories of the body with a soccer kick (legs), and with a volleyball strike (arms).}
\label{fig:movemes}
\vspace{-3mm}
\end{figure}

\begin{figure*}[!t]
\centering
\includegraphics[width=\linewidth]{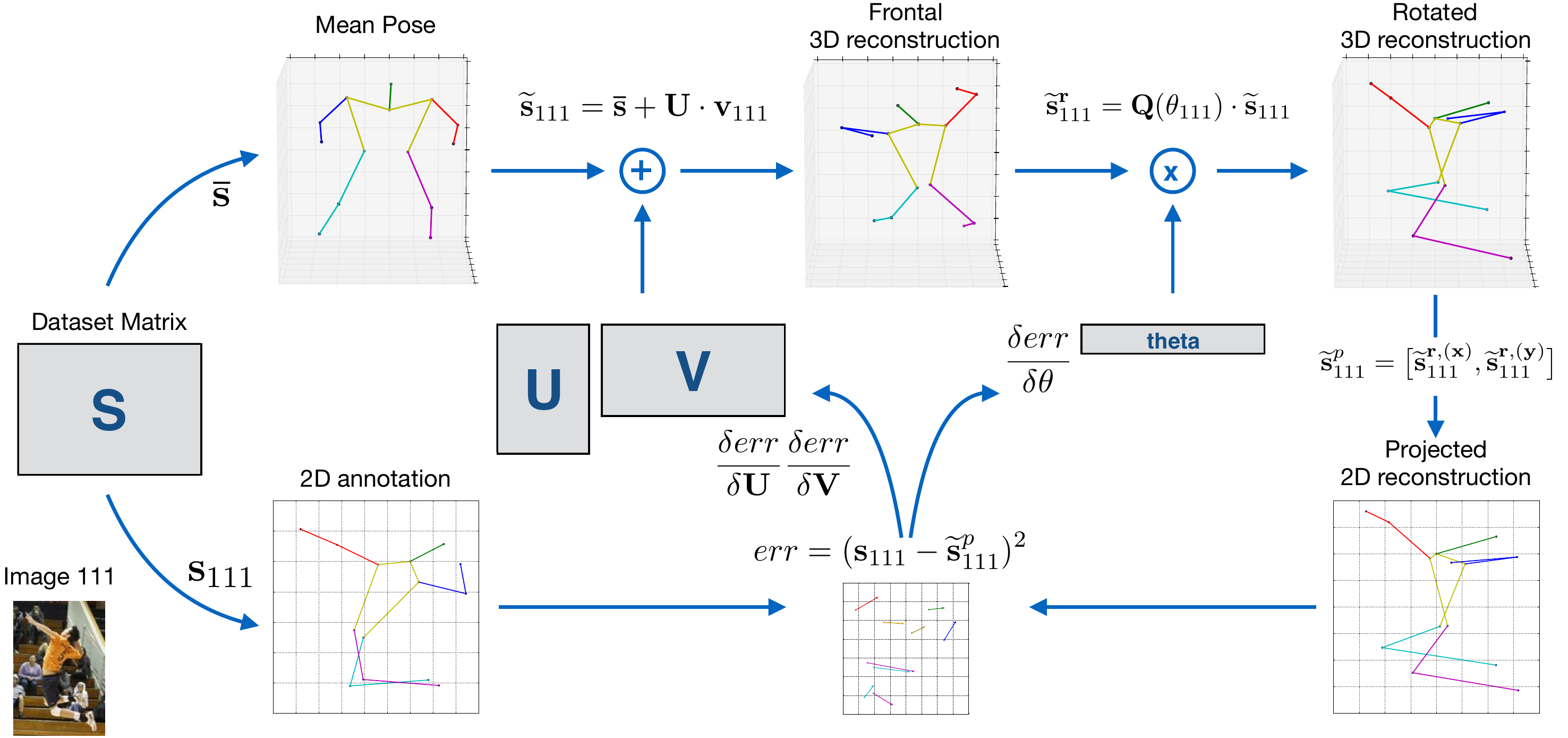}
\caption{ {\small \textbf{LFA3-D Method Pipeline}. Bases poses $\mathbf{U}$, coefficient matrix $\mathbf{V}$ and angles of view $\theta$ are initialized and updated through alternate stochastic gradient descent. Each iteration consists of the following steps: (1) a sparse linear combination of the current bases poses with coefficients from $\mathbf{V}$ is added to the dataset mean pose to obtain a frontal 3-D reconstruction of the true pose; (2) the 3-D reconstruction is rotated by the current estimate of the angle of view $\theta_j$ for that pose; (3) the 3-D pose is projected to the 2-D space where it is compared to the ground truth; (4) the gradient update step is computed to minimize the root mean square error \textit{wrt.} to quantities $\mathbf{U}$, $\mathbf{V}$ and $\theta$.}}
\label{fig:pipeline}
\end{figure*}

\subsubsection*{Clustered Singular Value Decomposition} \label{sec:bucketed_svd}
If the viewing angle of the training data is available, or a quantized approximation of it, then the basic latent factor model \eqref{eqn:basic} can be instantiated separately for different viewing angles, via:
\begin{eqnarray}
\mathbf{s}_j = \mathbf{U}(a_j)\cdot \mathbf{v_j} + \mathbf{\bar{s}}(a_j),\label{eqn:clustered}
\end{eqnarray}
where $a_j$ denotes the viewing angle cluster that example $j$ belongs to.  In other words, given $p$ clusters, we learn $p$ separate latent factor models, one per cluster.
Intuitively, we expect this method to suffer less conflation between changes in pose due to a viewing angle rotation and true pose deformation, and the more clusters, the less susceptible.
%This method is an improvement of the standard SVD and learns a set of basis poses that capture only the displacement due to motion and not rotation. 
The main drawbacks are that: (i) the learned bases representation is not global, and will not be consistent across the clusters since they are learned independently, and (ii) the amount of training data per model is reduced, which can yield a worse representation.

\subsection{Rotation-Invariant Latent Factor Model} \label{sec:lfa}
%Matrix factorization models map ground-truth poses and movemes to a joint latent space of lower dimensionality, such that their interactions can be modeled as inner products. 
Our goal is to develop a latent factor model that can learn a global representation of bases poses across different angles. For simplicity, we restrict ourselves to settings where there are only differences in the pan angle, and assume no variation in the tilt angle (i.e., all horizontal views).  
%In this framework, we seek a factorization that yields minimal reconstruction error across all the poses in the dataset and is able to marginalize the angle of view at which each pose is taken. 
To that end, we propose both a 2-D and a 3-D model which can be used depending on the quality and quantity of additional information available at training time. For some applications it may suffice to use the 2-D model, however the 3-D model is generally better able to intrinsically capture rotation-invariance. 

We first motivate some of the desirable properties:
\begin{itemize}
\item \textbf{Unsupervised} -- the bases discovery should not be limited to or dependent on images of specific classes of actions.
\item \textbf{Rotation Invariant} -- the learned bases should be composed of movements from a given canonical view (e.g., frontal) and be able to reconstruct poses oriented at any angle. The exact same pose may look different when observed from different camera angles; as such, it is important to disambiguate pose from viewing angle.
\item \textbf{Sparse} -- to encourage interpretability, the learned bases should be sparsely activated for any training instance. %, only a subset of which is used at a time to reconstruct specific poses. This will encourage the developement of shared basis across different activities. 
\item \textbf{Complementary} -- our method should be easy to integrate with other modeling approaches, and thus should implement an orthogonal extension of the basic latent factor modeling framework. %other algorithms by adding our spatial regularization terms. 
\end{itemize}

%\subsubsection*{Pose Representation}

%We can represent poses as relative coordinates with respect to a root joint, but this did not yield significant improvements. In the future we would like to use polar representation in order to enforce physical constraints such as joint angle limits \cite{akhter2015pose}. 
% \subsubsection*{Joint angle limits}
% Since we propose our problem as the solution to the gradient descent the cost function can be extended to incorporate regularization that ensures that joint angle limits of the human body are always respected. \cite{akhter2015pose}. Or better spatial regularization.

%The followig formalization of the problem is independent of the type of data available, later in the sections \ref{sec:lfa_2d}, \ref{sec:lfa_3d} we will detail the differences and advantages of each model separately. 
\subsubsection*{General Framework}
%Human pose is represented as a concatenation of 2-D or 3-D coordinates of the $d$ annotated joints ($d=14$ in our experiments).  For example, in the 3-D case, each training example takes the form: $\mathbf{s}_{j} = [\mathbf{x}_j^{T}, \mathbf{y}_j^{T}, \mathbf{z}_j^{T}]^{T} \in {\rm I\!R}^{3 \times d}$, so the total dimensionality of the data is $3d$. 
Our general framework aims to learn a latent factor matrix $\mathbf{U}$, containing the bases poses instantiated globally across all the training data; a coefficient matrix $\mathbf{V}$, whose columns correspond to the weights given to the bases poses to reconstruct all training instance; and a vector $\theta$, containing the angle of view of each training pose. 

We can thus model every training example as:
\begin{eqnarray}
\mathbf{s}_j = f(\mathbf{\bar{s}} + \mathbf{U}\cdot\mathbf{v}_j,\theta_j),\label{eqn:3d}
\end{eqnarray}
where $f(\cdot,\cdot)$ is a projection operator of the higher-dimensional model into the two-dimensional space. We train our model via:
\begin{align} 
    \mathbf{U},\mathbf{V},\mathbf{\theta} &= \argmin_{\mathbf{U},\mathbf{V},\mathbf{\theta}} \mathcal{L}(\mathbf{U}, \mathbf{V}, \mathbf{\theta}),\label{eq:cost_function}\\
    \mathcal{L}(\mathbf{U}, \mathbf{V}, \mathbf{\theta}) &= \mathcal{E}(\mathbf{U}, \mathbf{V}, \mathbf{\theta}) + \Omega(\mathbf{U},\mathbf{V},\theta),\label{eq:cost_function2}\\
    \mathcal{E}(\mathbf{U},\mathbf{V},\theta) &= \sum_j \left(\mathbf{s}_j - f(\mathbf{\bar{s}} + \mathbf{U}\cdot \mathbf{v}_j,\theta_j)\right)^2,\label{eq:cost_function3}
\end{align}
% \setlength{\arraycolsep}{0.0em}
% \begin{eqnarray}
% Z&{}={}&x_1 + x_2 + x_3 + x_4 + x_5 + x_6\nonumber\\
% &&+a + b\\
% &&+{}a + b\\
% &&{}+a + b\\
% &&{+}\:a + b
% \end{eqnarray}
% \setlength{\arraycolsep}{5pt}
where $\mathcal{E}$ is the squared reconstruction error over the training instances, and $\Omega$ is a model-specific regularizer. The projection operator $f$ and the regularizer $\Omega$ are specified separately for the 2-D and 3-D approach. This optimization problem is non-convex, and requires a reasonable initialization in order to converge to a good local optimum.\\

\subsubsection{2-D approach} \label{sec:lfa_2d}
The 2-D approach, uses the same approach as the clustered SVD baseline and, given a set of $p$ angle clusters, instantiates the projection operator as:
%\begin{equation} \label{eq:2d_lfa_cost_1}
%    \mathcal{E}(\mathbf{U}, \mathbf{V}, \mathbf{\theta}) = %\sum_{i} \sum_{j \in S^{(a)}} (S_{ij}(a) - U_i(a) V_{j}^{T}(a))^2\\
%\end{equation}
\begin{equation}
f(\mathbf{\bar{s}} + \mathbf{U}\cdot\mathbf{v}_j,\theta_j) = \mathbf{\bar{s}}(a_j) + \mathbf{U}(a_j)\cdot\mathbf{v}_j,\label{eqn:project_2d}
\end{equation}
$a_j$ denotes the cluster that $\theta_j$ belongs to, and a separate rank-$k$ $\mathbf{U}$ is learned for each viewing angle cluster. At this point, \eqref{eqn:project_2d} looks identical to \eqref{eqn:clustered}.  However, we encourage global consistency between the per-cluster models via the regularization terms:
\begin{equation}
     \Omega(\mathbf{U},\mathbf{V},\theta) = R_{reg}(\mathbf{U},\mathbf{V},\theta) + R_{spat}(\mathbf{U},\mathbf{V},\theta).\label{eqn:reg_2d}
\end{equation}
The first term in \eqref{eqn:reg_2d} is a standard regularizer used to prevent overfitting:
\begin{equation} \label{eq:2d_lfa_cost_2}
    R_{reg}(\mathbf{U}, \mathbf{V}, \mathbf{\theta}) = \sum_{a=1}^{p} \bigg[ \lambda_U \|\mathbf{U}(a)\|^2_{F} + \lambda_V\|\mathbf{V}(a)\|_1 \bigg].
\end{equation}
We wish to have sparse activations so we regularize $\mathbf{V}$ using L1 norm. Depending on the application, Sec.~\ref{sec:analysis}, we sometime enforce that $\mathbf{V}$ be non-negative for added interpretability.

The second term in \eqref{eqn:reg_2d} is the spatial regularizer that encourages (or in some cases enforces) consistency across the per-cluster models:
\begin{align} %\label{eq:2d_lfa_cost_3}
    R_{spat}(\mathbf{U}, \mathbf{V}, \mathbf{\theta}) &= \lambda_{spat} \sum_{a,a'} \kappa_{a,a'}\|\mathbf{U}^{(x)}(a) - \mathbf{U}^{(x)}(a')\|^2_{F}\label{eqn:reg_spatial1}\\
    &\ \ \ + \sum_{a,a'}\mathbf{1}\bigg(\mathbf{U}^{(y)}(a), \mathbf{U}^{(y)}(a')\bigg),\label{eqn:reg_spatial2}
\end{align}
$\mathbf{U}^{(x)}$ and $\mathbf{U}^{(y)}$ represent the $x$ and $y$ coordinate portion of the bases poses: e.g. $\mathbf{U}^{(x)} = [\mathbf{U}_{i,-}]$, $i \in X$, where $X$ is the set of indices corresponding to $x$ coordinates in the pose representation. Since we are only modeling variations in the pan angle, the $x$ coordinates can vary across different viewing angles, while the $y$ coordinates should remain constant. As such, the first term in $R_{spat}$, \eqref{eqn:reg_spatial1}, corresponds to encouraging the $\mathbf{U}^{(x)}(a)$ and $\mathbf{U}^{(x)}(a')$ of different clusters to be similar to each other (with $\kappa_{a,a'}$ controlling the degree of similarity), and the second term, \eqref{eqn:reg_spatial2}, is a $\{0,\infty\}$ indicator function that takes value 0 if the two arguments are identical, and value $\infty$ if they are not (i.e., it is a hard constraint). %\footnote{The exact components depend on the type of pose representation.}

In summary, the spatial regularization term is the main difference between the 2-D latent factor model and the clustered SVD baseline. Global consistency of the per-cluster models is obtained by encouraging similar values in the $x$ coordinates, and enforcing identical $y$ coordinates. In a sense, one can view spatial regularization as a form of multi-task regularization, which enables sharing statistical strength across the clusters. The main limitation of the 2-D model is that the spatial regularization does not incorporate more sophisticated geometric constraints, so the notion of consistency achieved may not align with the true underlying three-dimensional data.\\

%which varies based on the angle partition $a$. 

% \begin{equation} \label{eq:2d_lfa_cost}
%     \begin{cases}
%     \mathcal{E}(\mathbf{U}, \mathbf{V}, \mathbf{\theta}) &= \sum_{i} \sum_{j \in S^{(a)}} (S_{ij}(a) - U_i(a) V_{j}^{T}(a))^2\\
%     R_{reg}(\mathbf{U}, \mathbf{V}, \mathbf{\theta}) &= \\
%     = \sum_{a=1}^{p} \bigg[ \lambda \bigg(\frac{\|U^{(y)}\|^2}{p} +&\|U^{(x)}(a)\|^2 + \|V^T(a)\|^2 \bigg]\\
%     R_{spatial}(\mathbf{U}, \mathbf{V}, \mathbf{\theta}) &= \\
%     = \sum_{a=1}^{p} \bigg[\lambda \bigg(\sum_{a'}\kappa_{a,a'}&\|U^{(x)}(a) - U^{(x)}(a')\|^2\bigg)\bigg]
%     \end{cases}
% \end{equation}

%The constant $\kappa_{a,a'}$ denotes how similar we want the bases of two partitions to be, and can be set by the experimenter. In our experiments we used a simple procedure, where $\kappa_{a,a'} = \gamma$ is constant for  neighboring partitions $a$ and $a'$ and, and $\kappa_{a,a'} =0$ otherwise. This way, the influence of a basis pose on neighboring partitions decays exponentially with the distance between partitions.

\subsubsection{3-D approach} \label{sec:lfa_3d}
The 3-D model directly learns a three-dimensional representation of the underlying pose space, through a single and global $\mathbf{U} \in \Re^{3d \times k}$ that is inherently three-dimensional, and captures $k$ bases poses.
%\begin{equation} \label{eq:3d_lfa_cost_1}
%    \mathcal{E}(\mathbf{U}, \mathbf{V}, \mathbf{\theta}) = \sum_{i} (S_i - f(UV_{i}^T,\theta_i))^2\\
%\end{equation}

% \begin{equation} \label{eq:3d_lfa_cost_3}
%      R_{spatial}(\mathbf{U}, \mathbf{V}, \mathbf{\theta}) = \\
% \end{equation}
The projection operator is now defined as:
%where
\begin{equation} \label{eqn:3d_proj_operator}
f(\mathbf{\bar{s}} + \mathbf{U}\cdot\mathbf{v}_j,\theta_j) = \bigg[\mathbf{Q}(\theta_j)\bigg(\mathbf{\bar{s}} + \mathbf{U}\cdot\mathbf{v}_j\bigg)\bigg]^{(x,y)},
%\begin{cases}
%f(\mathbf{\bar{s}} + \mathbf{U}\cdot\mathbf{v}_j,\theta_j)^{(y)}&{}= \mathbf{\bar{s}}^{(y)}\mathbf{U}^{(y)}\cdot\mathbf{v}\\
%f(\mathbf{\bar{s}} + \mathbf{U}\cdot\mathbf{v}_j,\theta_j)^{(x)}&{}= \mathbf{Q}(\theta)\cdot\bigg(\bar{s}}^{(x)}+\mathbf{U}^{(x)}\cdot\mathbf{v}]^{(x)}
%\end{cases}
\end{equation}
where $\mathbf{Q}(\cdot)$ is the 3-D rotation matrix around the vertical axis:
\begin{equation}
\mathbf{Q}(\theta_j) = \left[\begin{array}{ccc}
\cos(\theta_j) & 0 & \sin(\theta_j)\\
0 & 1 & 0\\
-\sin(\theta_j) & 0 & \cos(\theta_j)
\end{array}\right],
\end{equation}
and the superscript $^{(x,y)}$ denotes the projection from the 3-D space of $\mathbf{U}$ to the 2-D space of the dataset annotations, obtained by indexing only the $x$ and $y$ coordinates (the underlying model provides $x$, $y$, and $z$ coordinates). The projection operator in~\eqref{eqn:3d_proj_operator} allows to compute the two-dimensional projection of any underlying three-dimensional pose at any viewing angle $\theta_j$ using standard geometric rules. Spatial regularization is no longer needed, because the rotation operator $\mathbf{Q}$ relates all the viewing angles to a common model, thus the regularizer assumes the standard form:
\begin{equation} \label{eq:3d_lfa_cost_2}
     \Omega(\mathbf{U},\mathbf{V},\theta) = \lambda_U\|\mathbf{U}\|^2_{F} + \lambda_V\|\mathbf{V}\|_1.
\end{equation}

In summary, the 3-D latent factor model improves upon the 2-D version by learning a global representation that is intrinsically three-dimensional and integrates domain knowledge of how the viewing angle affects pose via geometric projection rules. This results in a more robust method, that does not learn a separate model per viewing angle or rely on the spatial regularization to obtain consistency. The main drawback is that a more complex initialization will be required.

%Differently form the two-dimensional model we introduced in this case we do not need to add a spatial regularization to encode the rotation invariance but this model is intrinsically rotation invariant since the angle of view $\theta$ is treated as a latent variable and optimized along with U and V.

%A very appealing aspect is that we do not need supervision on the static images and poses that we use to learn our model, but we can just learn the model in an unsupervised way.

% Our method is based onto finding a decomposition of human motion from static 2d images, into elemental recognizable movements, that we can then use to recombine and generate a new action or to discriminate what action a specific pose is closest to.

%\subsubsection*{Sparse Coding}
%similar strategy used by olhausen and Field in \cite{olshausen1997sparse}, where an overcomplete basis set is used to produce a sparse representation of an image. We put basis in competition with each other by using l1 norm on both U and V matrix and using iterative soft thresholding algorithm \cite{daubechies2004iterative,beck2009fast}.

\subsection{Training Details}

\subsubsection*{Initialization}
Our approaches require an initial guess of the viewing angle for each training instance, and the bases poses $\mathbf{U}$. For angle initialization, we show in our experiments (Sec.~\ref{exp:angles}) that we only need a fairly coarse prediction of the viewing angle (e.g., into quadrants). The 2-D latent factor model bases poses $\mathbf{U}$ are initialized uniformly between -1 and 1, while for the 3-D model we use an off-the-shelf pose estimator \cite{akhter2015pose} and initialize $\mathbf{U}$ as the left singular vectors of the mean centered 3-D pose data, obtained through SVD.

%requires the optimization to be done in the three-dimensional space, so we initalized it using the SVD of the noisy 3-D poses estimated with \cite{akhter2015pose}. 

%We show in the experiments how to specify a reasonable initialization for training an effective model.

\subsubsection*{Optimization}
For both models, we optimize Eq.~\eqref{eq:cost_function} using alternating stochastic gradient descent, divided in two phases:
\begin{itemize}
\item Representation Update: we employ standard stochastic gradient descent to update $\mathbf{U}$ and $\mathbf{V}$ while keeping $\theta$ fixed. For the 3-D model, this involves computing how the training data (which are two-dimensional projections) induce a gradient on $\mathbf{U}$ and $\mathbf{V}$ through the rotation $\mathbf{Q}$. Because we employ an L1 regularization penalty, we use the standard soft-thresholding technique  \cite{beck2009fast}.
\item Angle Update: Once the optimal $\mathbf{U}$ and $\mathbf{V}$ are fixed, we employ standard stochastic gradient descent to update $\theta$.
\end{itemize}
Fig.~\ref{fig:pipeline} provides an overview of the steps for the 3-D approach.\vspace{2mm}

\subsubsection*{Convergence and Learning Rates}
Three training epochs of 10000 iterations are usually sufficient for convergence to a good local minimum. Typical values of the learning rate are \num{1e-4} for $\mathbf{U}$ and $\mathbf{V}$ and \num{1e-6} for $\theta$. We use a smaller step size in the update of $\theta$, since the curvature of the objective function~\eqref{eq:cost_function} w.r.t. $\theta$ is higher than w.r.t. $\mathbf{U}$ and $\mathbf{V}$.

\begin{figure*}[t!]
\begin{center}
\includegraphics[width=\linewidth]{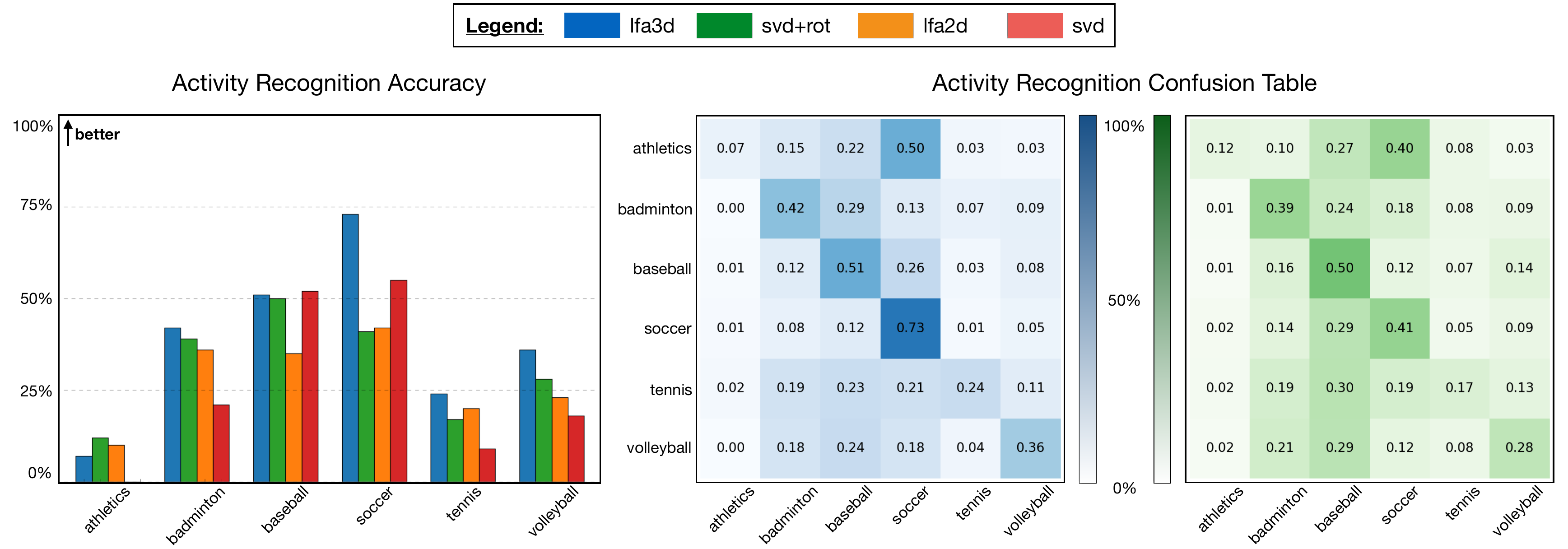}
\end{center}
\caption{ {\small \textbf{Activity Recognition Performance}. (Left) The activity classification accuracy across the sports in LSP for the following methods: ``svd'' -- baseline, ``svd+rot'' -- clustered version of the baseline, ``lfa2d'' -- 2-D latent factor model with spatial regularization, ``lfa3d'' -- full 3-D latent factor model. (Right) The confusion tables for the best two performing methods, ``lfa3d'' and ``svd+rot''. Full details in Sec.~\ref{exp:classification}.}}
\label{fig:exp_classification}
\vspace{-3mm}
\end{figure*}

%%%%%%%%%%%%%%%%%%%%%%%%%%%%%%%%%%%%%%%%%%%%%%%%%%%%%%%%%%%%%%%%%%%%%%%%%%%%%%%%%%%%%%%%%%
%%%%%%%%%%%%%%%%%%%%%%%%%%%%%%%%%%%%%%%%%%%%%%%%%%%%%%%%%%%%%%%%%%%%%%%%%%%%%%%%%%%%%%%%%%
%% Experiments
%%%%%%%%%%%%%%%%%%%%%%%%%%%%%%%%%%%%%%%%%%%%%%%%%%%%%%%%%%%%%%%%%%%%%%%%%%%%%%%%%%%%%%%%%%
%%%%%%%%%%%%%%%%%%%%%%%%%%%%%%%%%%%%%%%%%%%%%%%%%%%%%%%%%%%%%%%%%%%%%%%%%%%%%%%%%%%%%%%%%%

%%%%%%%%%%%%%%%%%%%%%%%%%%%%%%%%%%%%%%%%%%%%%%%%%%%%%%%%%%%%%%%%%%%%%%%%%%%
%%%% EXPERIMENTS %%%%%%%%%%%%%%%%%%%%%%%%%%%%%%%%%%%%%%%%%%%%%%%%%%%%%%%%%%
\section{Experiments} \label{sec:experiments}

%%%% LSP DATASET %%%%%%%%%%%%%%%%%%%%%%%%%%%%%%%%%%%%%%%%%%%%%%%%%%%%%%%%%%
\subsection{Dataset and Additional Annotations} \label{sec:datasets}
We use the Leeds Sports Dataset (LSP) \cite{Johnson10} for our experiments. LSP is composed of 2000 images containing a single person performing one of eight sports (Athletics, Badminton, Baseball, Gymnastics, Parkour, Soccer, Tennis, Volleyball) annotated with the x,y location and a visibility flag for 14 joints of the human body. Example images and annotations are shown in Fig.~\ref{fig:introduction}, \ref{fig:movemes} and~\ref{fig:tsne}. 
Sports activities are particularly well suited for this study, as they present characteristic motions that share trajectories of parts of the body, that allow investigating basis pose sharing across sports. As part of preprocessing, we normalize all the poses in the dataset by modifying each bone to have the average bone length computed over all the training instances \cite{fan2014pose}.
We discard ``Gymnastics'' and ``Parkour'' from our analysis because they have few examples and the class poses do not vary exclusively along the pan angle (but appear in very unconventional views, i.e. upside-down and horizontal), violating the assumption in Sec.~\ref{sec:lfa}. Generalizing the framework, to incorporate a wider variability of the viewing angles, is an interesting future direction.

We collected high-quality viewing angle annotations for each pose in LSP. Although these annotations are not necessary for training, we use them to demonstrate the robustness of our model to poor angle initialization, and that it can in fact recover the ground truth value, see Sec.~\ref{exp:angles}. Three annotators evaluated each image and were instructed to provide the direction at which the torso was facing\footnote{The angle annotations for LSP, annotator agreement statistics, and details about the Amazon Mechanical Turk GUI are available at the project page~\cite{ronchi-movemes}.}. The standard deviation in the reported angle of view averaged over the whole dataset is 12 degrees, and more than half of the images have a deviation of less than 10 degrees, showing a very high annotator agreement for the task.

%%%% ANALYSIS %%%%%%%%%%%%%%%%%%%%%%%%%%%%%%%%%%%%%%%%%%%%%%%%%%%%%%%%%%%%%
\subsection{Empirical Results} \label{sec:analysis}

We analyze the flexibility and usefulness of the proposed model in a variety of application domains and experiments. In particular, we evaluate (i) the performance of the learned representation for supervised learning tasks such as activity classification; (ii) whether the learned representation captures enough semantics for meaningful manifold traversal and visualization; and (iii) the robustness to initialization and the generalization error. Collectively, results suggest that our approach is effective at capturing rotation invariant semantics of the underlying data.\\

%%%% ACTIVITY RECOGNITION %%%%%%%%%%%%%%%%%%%%%%%%%%%%%%%%%%%%%%%%%%%%%%%%
\subsubsection{Activity Recognition} \label{exp:classification} 
The matrix $\mathbf{V}$ describes each pose in the dataset as a linear combination of the learned latent factors, Sec.~\ref{sec:notation}. Thus, $\mathbf{v}_j$ can be interpreted as a semantically more meaningful feature representation for $j$-th data point. For instance, if a lower body basis pose (e.g. Fig.~\ref{fig:sequences} top row) has a high weight, the reconstructed pose is very likely to represent a movement from an activity related to running, or kicking.

A natural way to test the effectiveness of the learned representation is to use it for supervised learning tasks. To that end, we used the coefficients in $\mathbf{V}$ as input features for classifying the sport categories in LSP. 

Fig.~\ref{fig:exp_classification} shows the results obtained from five-fold cross validation. The proposed 3-D latent factor model (``lfa3d'') outperforms all other methods by an average accuracy of about $11\%$. %The second best method overall is the clustered SVD baseline (``svd+rot''), which benefits from learning different basis poses for each angle of view cluster with a $5\%$ average improvement over the ``svd'' baseline.  
The 2-D model (``lfa2d'') performs slightly worse than the clustered SVD baseline (``svd+rot''), but both show more than a $5\%$ average improvement over the ``svd'' baseline.
The two most challenging activities are ``athletics'', which does not posses characterizing movements; and ``tennis'', whose movemes are shared and thus confused with multiple other sports, ``badminton'' and ``baseball'' above all. We also report the full classification confusion tables in Fig.~\ref{fig:exp_classification}.
%These results suggest that the learned representation from the 3-D model improves the usability of latent factor methods for supervised classification settings that benefit from capturing rotational invariance.
Note that only the weights of the latent factors reconstructing a pose are being used to discriminate between the activities, without the aid of visual cues from the image. It is thus surprising that ``lfa3d'' achieves an average $39\%$ accuracy, when a random guess would merely give $16.7\%$. Finally, the obtained feature representation is complementary to other representations, such as the hidden layer activations of a convolutional neural network~\cite{krizhevsky2012imagenet}, and we wish to investigate in future work the performance obtained by their combination.\\

%These results suggest that latent factor methods for supervised classification settings benefit from capturing rotational invariance.\\

%%%% ACTION DYNAMICS INFERENCE %%%%%%%%%%%%%%%%%%%%%%%%%%%%%%%%%%%%%%%%%%%%
\begin{figure*}[t!]
\includegraphics[width=\linewidth]{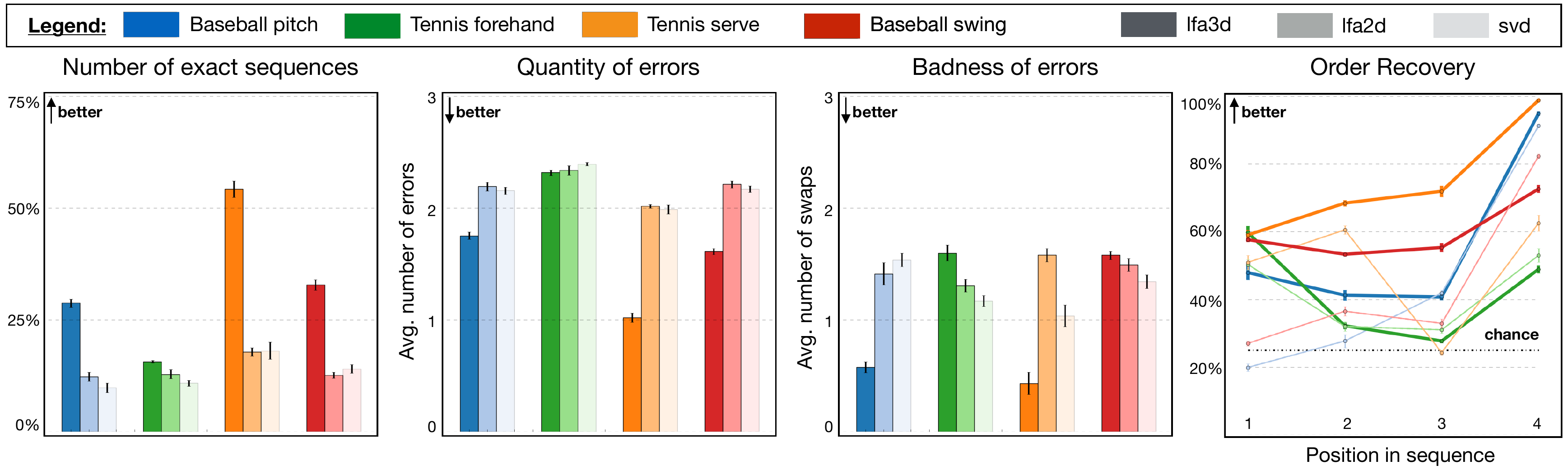}
\begin{tabular}{m{1.8cm}m{4cm}m{4cm}m{4cm}m{1.8cm}}
& (1) & (2) & (3) & (4)
\end{tabular}
\caption{ {\small \textbf{Action Dynamics Inference Performance}. We compare the methods ``svd'', ``lfa2d'', and ``lfa3d'' in the task of reordering shuffled sequences of images sampled from four different sport actions. The color scheme represents actions, the methods are plotted with a different transparency value. The performance is described in terms of: (1) number of sequences exactly reordered; (2) average number of errors contained in a sequence; (3) average number of swaps needed to obtain the correct sequence; (4) accuracy per position in the sequence -- shown only for the best two methods (``lfa3d'' - dark marker, ``lfa2d'' - light marker). Example sequences in Tab.~\ref{tab:sequence_reordering}. Full details in Sec.~\ref{exp:reordering}.}}
\label{fig:exp_dynamics}
\vspace{-3mm}
\end{figure*}

\subsubsection{Action Dynamics Inference \& Manifold Traversal} \label{exp:reordering}
Every pose in the training set belongs to a movement of the body corresponding to a complex trajectory in the manifold of human motion. If the latent factor model captures the semantics of the data, then poses that occur in chronological order within a given action should lie in a monotonic sequence within the learned space. A quantitative measure of the quality of the representation can be obtained by observing how well the order of poses belonging to a same action is preserved. One straightforward way to find the sequence in which a set of poses lies in the manifold, is to look at the coefficient of their projection along the ``total least squares'' line fit \cite{de1998introduction} of the corresponding columns in the matrix $\mathbf{V}$. In other words, we are computing a linear traversal through the representation space.
%we attempt to locally approximate these curves by a linear interpolation of the learned movemes. A way of quantitatively analyzing how good is the obtained approximation, is to observe how well is the order of sequences of poses which belong to a same action preserved. 
Furthermore, this ordering should hold regardless of the angle of view of the input instances.
%, in order to claim that the learned representation is rotation invariant.

In this experiment, we shuffled 1000 sequences of four images for four sport actions (``baseball pitch'', ``tennis forehand'', ``tennis serve'', ``baseball swing''), and verified how precisely could the underlying chronological sequence be recovered.
The analysis is repeated five times to obtain standard deviations, and performance is measured in terms of three metrics: (1) what percentage of the 1000 sequences is exactly reordered; (2) how many poses are wrongly positioned; and (3) how bad are the reordering mistakes, computed as the number of swaps necessary to correct a sequence. 

%For futher insights, we also show the accuracy of correctly identifying a specific position within a sequence for all the actions, Fig.~\ref{fig:exp_dynamics}-(4).

Fig.~\ref{fig:exp_dynamics} shows the results for the latent factor models ``lfa2d'', ``lfa3d'' and for the ``svd'' baseline. It is not possible to study the performance of the clustered baseline ``svd+rot'' since it does not learn a global matrix $\mathbf{U}$, thus the coefficients in $\mathbf{V}$ are not comparable across different viewing angles.

The ``lfa3d'' model has significantly better outcomes compared to ``lfa2d'' and ``svd'', which perform similarly. Specifically, ``lfa3d'' correctly reorders more than twice the sequences overall (1314 against 555 of ``lfa2d'') averages 1.6 errors, and is the only algorithm to require an average number of swaps smaller than 1. Fig.~\ref{fig:exp_dynamics}-(4) shows the per-position accuracy.% for ``tennis forehand'', 
%one example test sequence for ``tennis forehand'', 
%which is the hardest motion to order correctly due to significant ambiguity of the central poses of the sequence. %, and that typically the central poses of a sequence are the hardest to position accurately.

An example sequence for ``tennis serve'' is shown in Tab.~\ref{tab:sequence_reordering}. Only the ``lfa3d'' method recovers the order correctly; note how the images are all taken from different viewing angles.\\

%%%% MOVEME PURITY %%%%%%%%%%%%%%%%%%%%%%%%%%%%%%%%%%%%%%%%%%%%%%%%%%%%%%%
\subsubsection{Moveme Visualization} \label{exp:moveme_purity}
The ``lfa3d'' method can be used to recover and synthesize realistic human motions from static joint locations in images. The underlying idea, is that models of human motion can be successfully learned from observations of poses of people performing various actions, as opposed to deriving mathematical principles which define control laws (e.g. inverse kinematics).

The most significant movemes contained in the training set are captured by the bases poses matrix $\mathbf{U}$ and encoded in the form of a displacement from the mean pose. Each column of $\mathbf{U}$ corresponds to a latent factor that describes some of the movement variability present in the data.

\begin{figure*}[t!]
\begin{center}
\includegraphics[width=\linewidth]{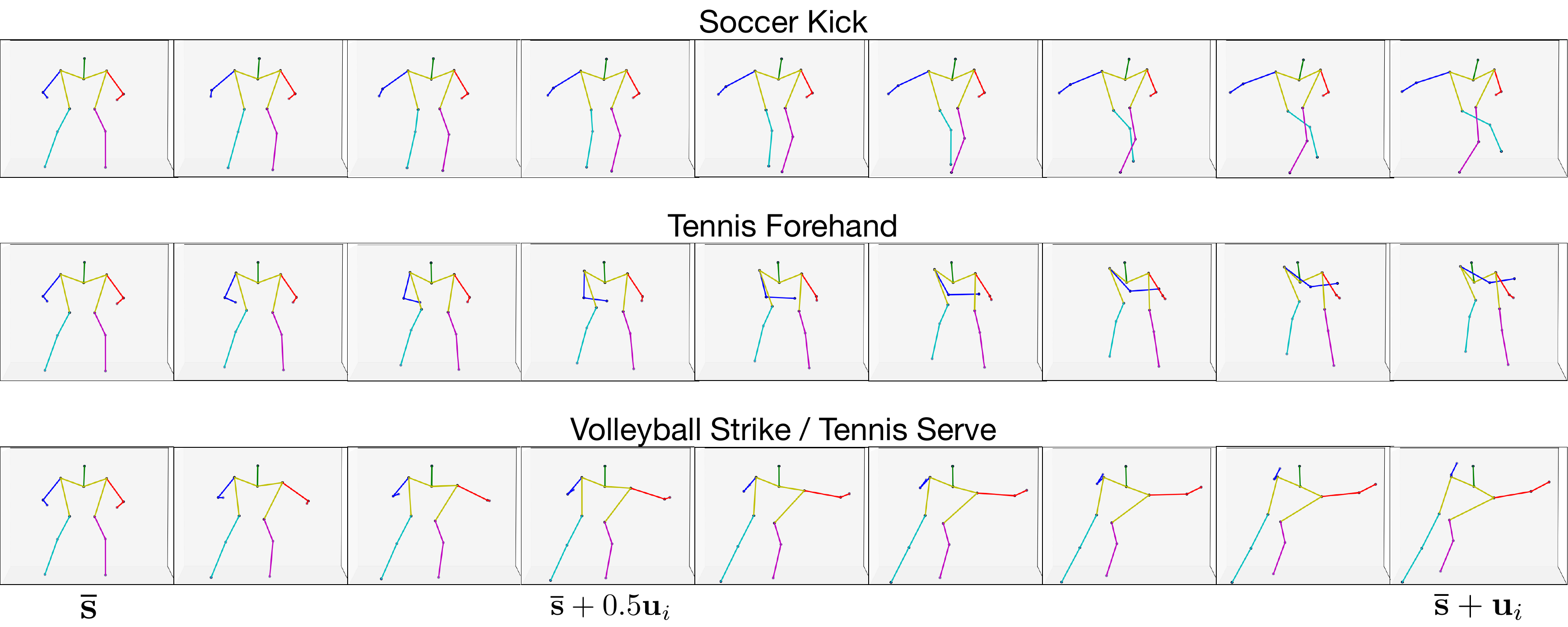}
\end{center}
\caption{ {\small \textbf{Learned Movemes Visualization}. Three latent factors, encoding movemes, from the learned bases poses matrix $\mathbf{U}$; two are easily interpretable (``soccer kick'', ``tennis forehand'') and one is not as well defined (``volleyball strike / tennis serve''). The sequences are obtained by adding an increasing fraction of the basis to the mean pose of the dataset and differentiate very clearly, as early as 30$\%$ of the final movement, as visible in the second column. Full details in Sec.~\ref{exp:moveme_purity}.}}
\label{fig:sequences}
\vspace{-3mm}
\end{figure*}

Fig.~\ref{fig:sequences} reports the motion described by three latent factors: the rows show the pose obtained by adding an increasing portion of the learned moveme (from $30\%$ - second column, to $100\%$ - last column) to the mean pose of the data (first column). Two are easily interpretable, ``soccer kick'' and ``tennis forehand'', while one is not as well defined, ``volleyball strike / tennis serve''. The movemes differentiate very quickly, as early as $30\%$ of the final movement is added.

We verify empirically that two parameters mainly affect the correspondence between an action and a latent factor (moveme purity): the number of latent factors, and the constraints put on the coefficients of $\mathbf{V}$. We obtain the best visualizations by approximately matching the number of latent factors with the number of recognizable actions contained in the dataset (10 for this experiment), and constraining the coefficients of V to be between 0 and 1.\\ 

% which results in the capability of reconstructing very different poses in a sparse way and using small weights in $\mathbf{V}$.

%This encourages the developement of latent factors that individually are capable of approximating poses contained in certain movements, without relying on other latent factors.\\

%as the number of recognizable actions contained in the dataset. Furthermore, constraining the V parameters to be positive and less than one enforces the basis pose to develop in a way that they can differentiate to describe the whole span of specific types of movements in the dataset. It is a desirable property that they differentiate quickly and clearly.

% There is a tradeoff between sparsity of the learned basis poses (useful for applications such as activity classification and action dynamics inference) and the quality of the visualization of the movement. This is not necessarily a drawback of our method, but instead illustrates its flexibility in terms of tuning parameters that allow it to serve different purposes.\\

% If an action is described best by the presence of a certain set of movemes we can try to predict and infer the evolution of a certain pose once we identify that it lies in a specific part of the manifold we identified; or to simply transform an action into another one: we can imagine how a pose that is soccer like, could be transormed into a baseball like pose by changing the parameters of the movemes in such a way that brings it in a different part of the learned manifold.

\begin{table}[!t]
\renewcommand{\arraystretch}{1}
\caption{We use the coefficients in $\mathbf{V}$ to order chronologically four images sampled from a tennis serve. For each method, we report the number of images out of position and swaps necessary to obtain the correct order.}
\label{tab:sequence_reordering}
\centering
\begin{tabular}{U V U U}
\Xhline{2.5\arrayrulewidth}
\bfseries Method & \bfseries Reordered Sequence & \bfseries Errors & \bfseries Swaps\\
\hline\hline
& & &\\
lfa3d & \fcolorbox{white}{yellow}{\includegraphics[width=0.85cm,height=1.55cm]{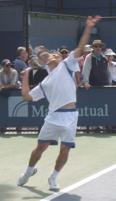}} \fcolorbox{white}{yellow}{\includegraphics[width=0.85cm,height=1.55cm]{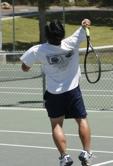}} \fcolorbox{white}{yellow}{\includegraphics[width=0.85cm,height=1.55cm]{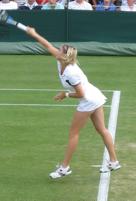}} \fcolorbox{white}{yellow}{\includegraphics[width=0.85cm,height=1.55cm]{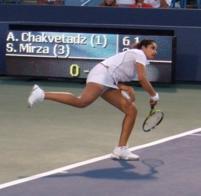}} & 0 & 0\\
lfa2d & \fcolorbox{white}{yellow}{\includegraphics[width=0.85cm,height=1.55cm]{im1655.jpg}} \fcolorbox{white}{yellow}{\includegraphics[width=0.85cm,height=1.55cm]{im1741.jpg}} \fcolorbox{red}{yellow}{\includegraphics[width=0.85cm,height=1.55cm]{im0142.jpg}} \fcolorbox{red}{yellow}{\includegraphics[width=0.85cm,height=1.55cm]{im1950.jpg}} & 2 &  1 \\
svd & \fcolorbox{red}{yellow}{\includegraphics[width=0.85cm,height=1.55cm]{im1950.jpg}} \fcolorbox{white}{yellow}{\includegraphics[width=0.85cm,height=1.55cm]{im1741.jpg}} \fcolorbox{red}{yellow}{\includegraphics[width=0.85cm,height=1.55cm]{im1655.jpg}} \fcolorbox{white}{yellow}{\includegraphics[width=0.85cm,height=1.55cm]{im0142.jpg}} & 2 & 3 \\
\hline
\end{tabular}
\vspace{-5mm}
\end{table}

%%%% ANGLE RECONSTRUCTION %%%%%%%%%%%%%%%%%%%%%%%%%%%%%%%%%%%%%%%%%%%%
\subsubsection{Angle Recovery} \label{exp:angles}
The ``lfa3d'' method learns a rotation invariant representation by treating the angle of view of each pose as a variable which is optimized through gradient descent (Sec.~\ref{sec:lfa_3d} and Fig.~\ref{fig:pipeline}), and requires an initial guess for each training instance. We investigate how sensitive is the model to initialization, and how close is the recovered angle of view to the ground truth.
Fig.~\ref{fig:other_exp}(a) shows the Root Mean Squared Error (RMSE) and cosine similarity with ground truth, for three initialization methods: (1) ``random'', between 0 and $2\pi$; (2) ``coarse'', coarsening into discrete buckets (e.g., 4 clusters indicates that we only know the viewing angle quadrant during initialization); and (3) ``ground-truth''.

%Going from ``random'' to ``ground-truth'' initialization provides almost a $20\%$ reduction in the RMSE, from 23.1 to 18.7, which is independent of the number of clusters in which we partition the poses. 
As the number of clusters increases, we see that performance remains constant for ``random'' and ``ground truth'', while both evaluation metrics improve significantly for ``coarse'' initialization. For instance, using just four clusters, ``coarse'' initialization obtains almost minimal RMSE and perfect cosine similarity. These results suggest that using very simple heuristics to predict the viewing angle quadrant of a pose is sufficient to obtain optimal performance.\\

\subsubsection{Generalization Behaviour} \label{exp:generalization}
%The latent factor matrix $\mathbf{U}$ captures the most significant movements of the human body contained in a dataset of static poses.
A desirable property of the obtained model is to be able to reconstruct with low error poses that are not contained in the training set, so the representation is not tied uniquely to the specific image collection it was learned from. To verify the generalization quality of the learned bases poses we trained the ``lfa3d'' model on a subset of the dataset and measured the RMSE on the remaining part, for an increasingly larger portion of the data. We repeated the experiment five times to obtain standard deviations.

\begin{figure}
\centering
\begin{tabular}{cc}
\hspace{-4mm}\includegraphics[height=0.4\linewidth]{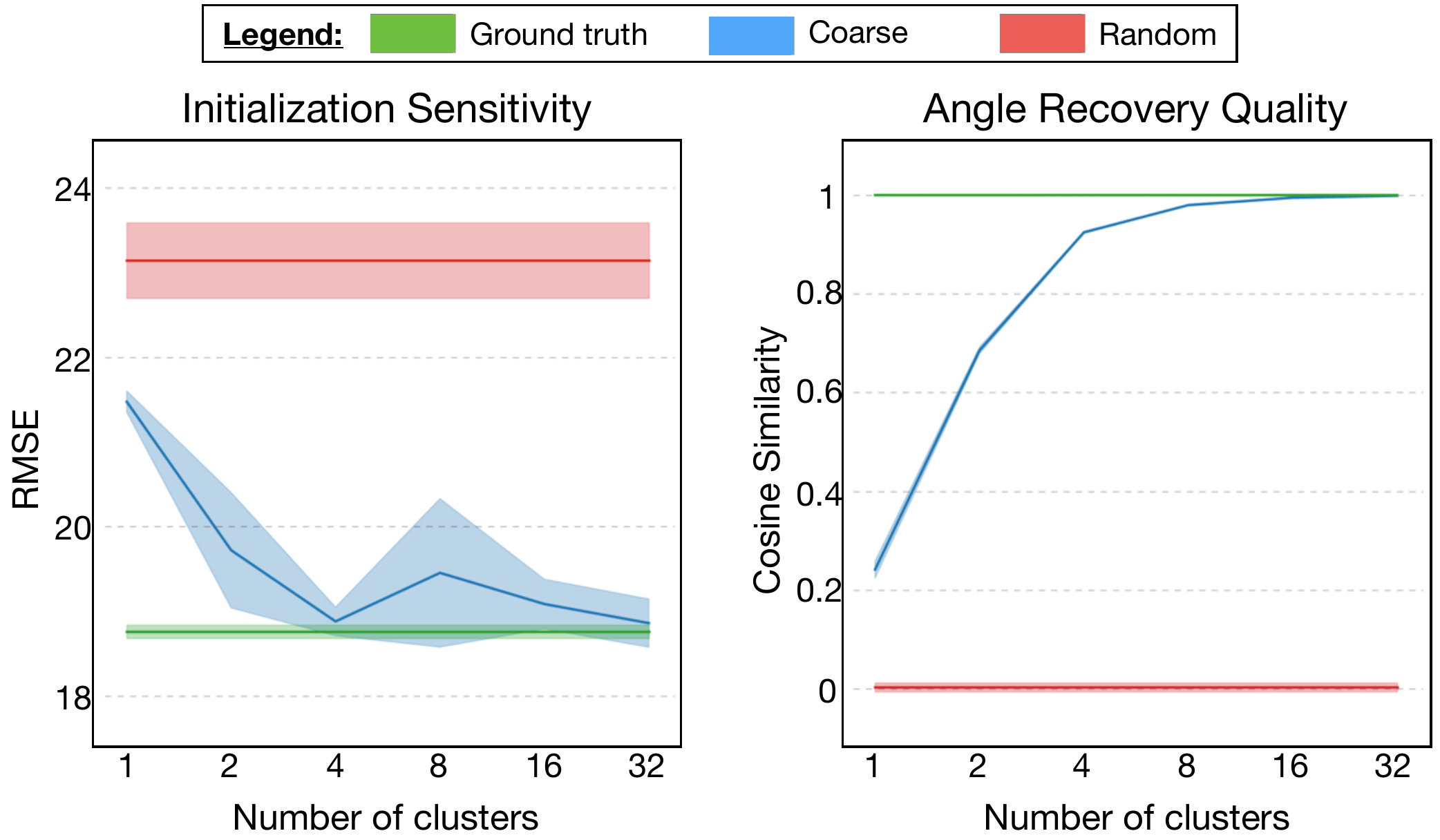} & \hspace{-3mm}\includegraphics[height=0.4\linewidth]{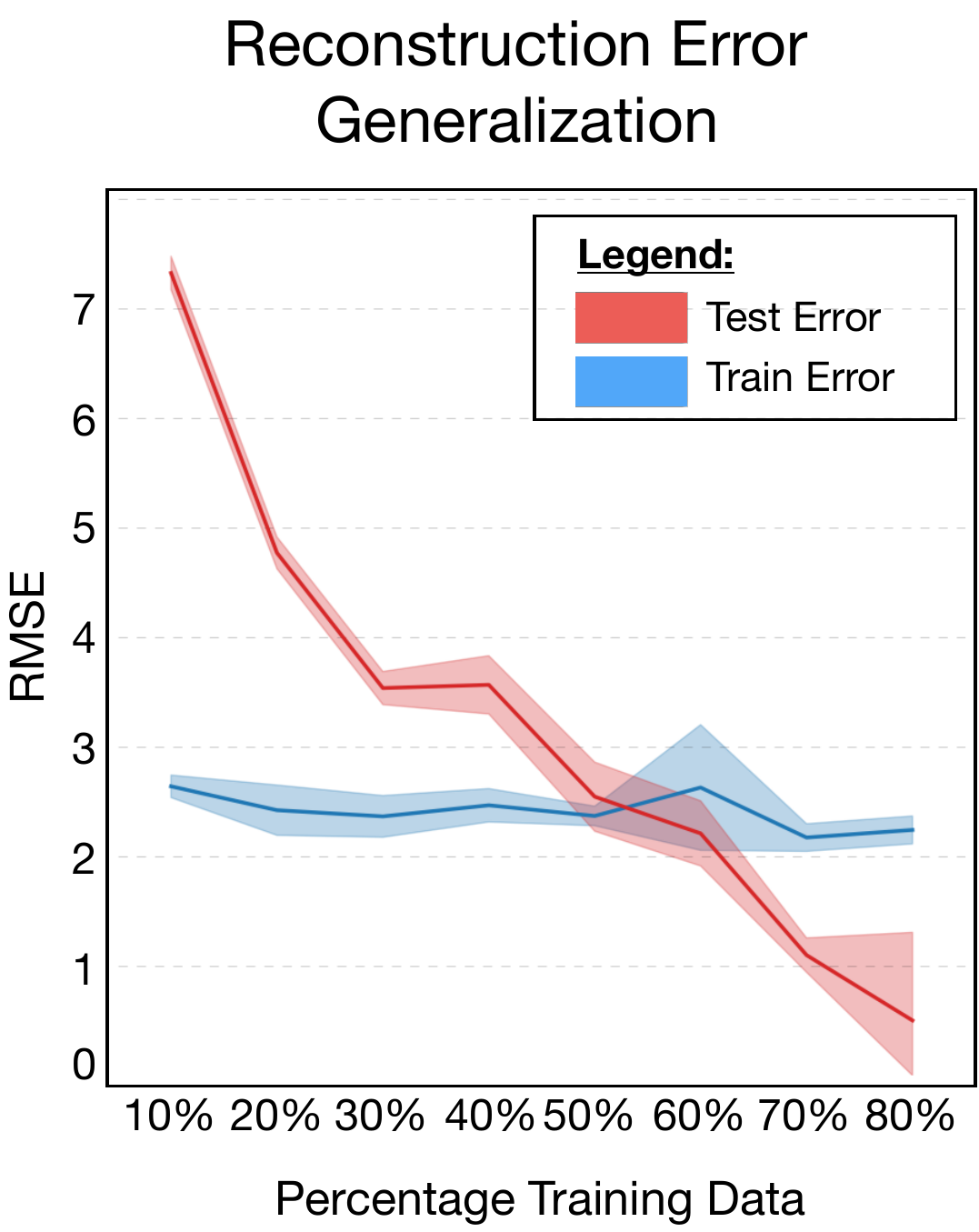}\\
(a) & (b)
\end{tabular}
\caption{ {\small \textbf{(a) Angle Recovery and (b) Generalization Performance}. (a) Sensitivity \textit{wrt.} the initial value of the angle of view of the training poses of (Left) the Root Mean Squared Error and (Right) the Cosine Similarity between the learned and ground truth angles. A coarse initialization, within the correct quadrant of the true value, yields performances similar to ground truth. (b) The reconstruction error for poses not contained in the training set \textit{wrt.} the percentage of data used in the training set. Full details in Sec.~\ref{exp:angles}, and Sec.~\ref{exp:generalization}.}}
\label{fig:other_exp}
\end{figure}

As reported in Fig.~\ref{fig:other_exp}(b), the RMSE over the training set is approximately constant, while the test set RMSE decreases significantly when going from $10\%$ to $80\%$ of the data used in training. This indicates that the learned latent factors can successfully reconstruct poses of unseen data.\\

\begin{figure*}[t!]
\begin{center}
\includegraphics[width=\linewidth]{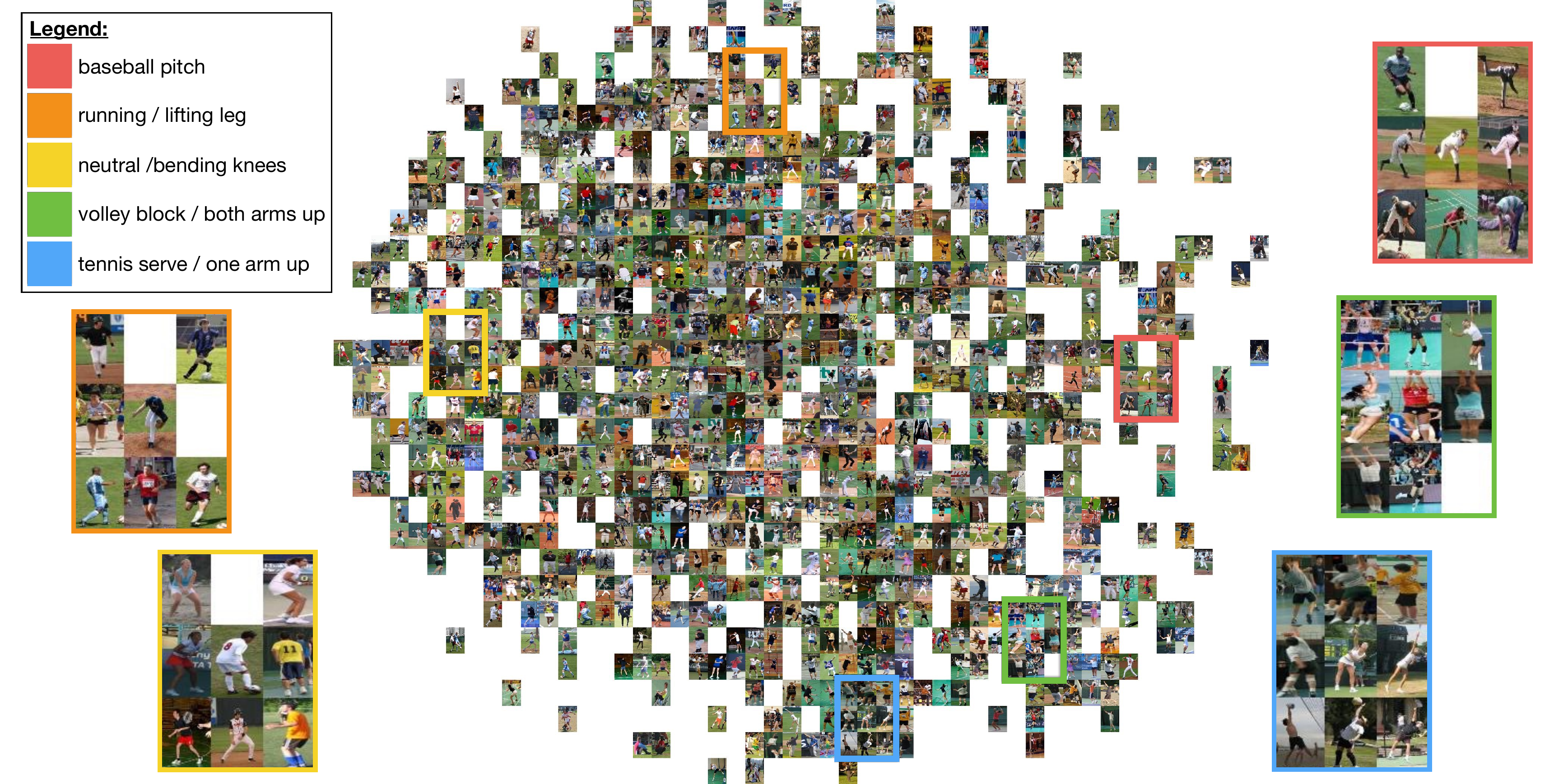}
\end{center}
\caption{ {\small \textbf{Human Motion Manifold Visualization}. t-SNE embedding of the poses contained in the LSP dataset. Images, instead of poses, are shown for interpretability purpose. The type of body movement, and the influence of the learned bases poses determine the location in the manifold:``tennis serve'' and ``volleyball block'' appear close in the manifold, while ``running'' is at the opposite end of the embedding. The angle of view does not affect the location in the manifold, as nearby poses may have very different angle of view. Full details in Sec.~\ref{exp:tsne}.}}
\label{fig:tsne}
\vspace{-3mm}
\end{figure*}

%%%% MANIFOLD VISUALIZATION %%%%%%%%%%%%%%%%%%%%%%%%%%%%%%%%%%%%%%%%%%%%
\subsubsection{Manifold Visualization} \label{exp:tsne}

Fig.~\ref{fig:tsne} visualizes an embedding of the manifold of human motion learned with the ``lfa3d'' method. Each pose in LSP is mapped in the human motion space through the coefficients of the corresponding column of $\mathbf{V}$ and then projected in two-dimensions using t-SNE \cite{van2008visualizing}.

Poses describing similar movements are mapped to nearby positions and form consistent clusters, whose relative distance depends on which latent factors are used to reconstruct the contained poses. Upper body movements are mapped closely in the lower right corner, while lower body movements appear at the opposite end of the embedding. The mapping in the manifold is not affected by the direction each pose is facing, as nearby elements may have very different angle of view, confirming that the learned representation is rotation invariant.
\begin{figure}[!t]
\centering
\includegraphics[width=\linewidth]{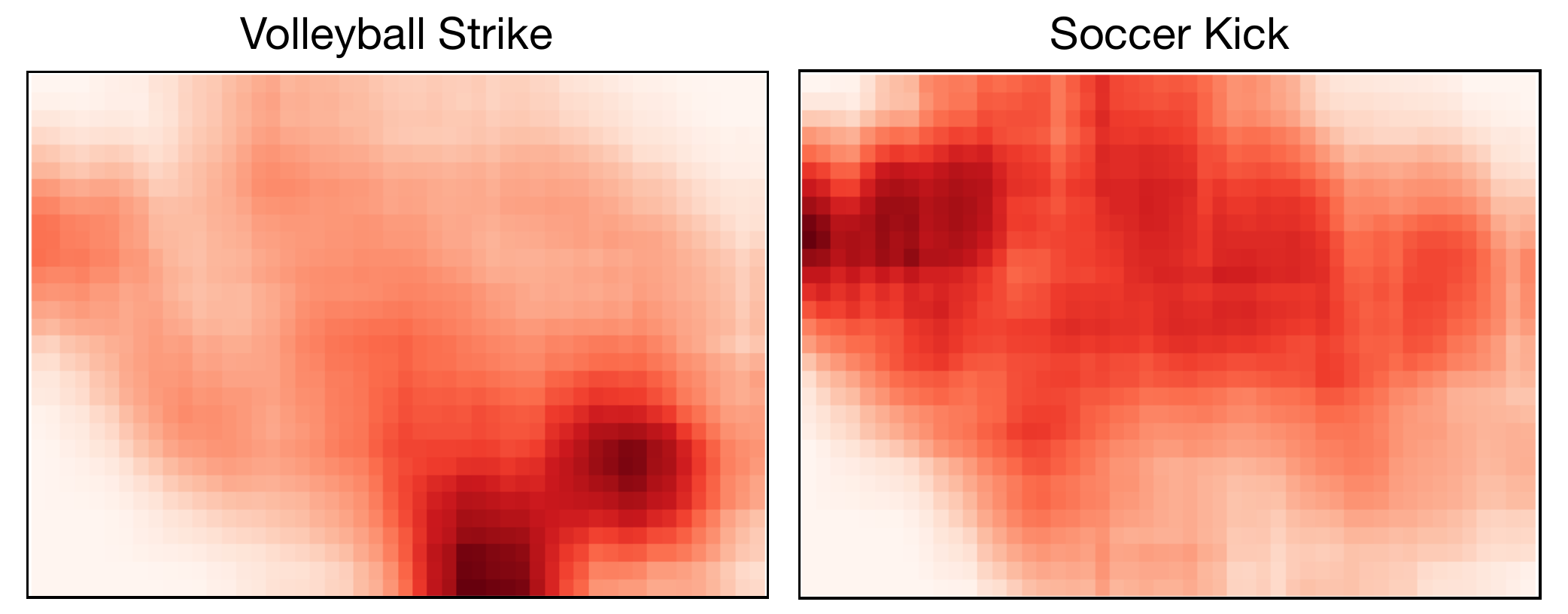}
\caption{{\small \textbf{Learned Movemes Heat-maps}. Activation strength of the learned ``volley strike'' and ``soccer kick'' bases poses from Fig.~\ref{fig:sequences} (third and first row) in the t-SNE embedding. The heat-maps are consistent with Fig.~\ref{fig:tsne} in which movements of the upper and lower body are respectively mapped to the low-right and high-left corner.}}
\label{fig:heatmaps}
\vspace{-3mm}
\end{figure}
%To reinforce the qualitative results of Fig.~\ref{fig:sequences} and~\ref{fig:tsne} we analyze the strength of the activation of the learned latent factors within the embedding obtained with t-SNE.
In Fig.~\ref{fig:heatmaps}, we show the heatmaps obtained from the activations of two latent factors from Fig.~\ref{fig:sequences}, overlaid on top of the t-SNE mapping of Fig.~\ref{fig:tsne}.
To compute the heatmaps, we extract the coefficients for the ``soccer kick'' and ``volleyball strike'' latent factors from each column of $\mathbf{V}$ corresponding to a location in the embedding, and plot their value after normalization\footnote{To better depict the high-level trends, we enhance the contrast using a power of 1.5 and employ Gaussian smoothing.}. \\

Clearly, the epicentrum of the ``volleyball strike'' basis pose is located where volleyball-like poses appear in the t-SNE plot (lower-right corner). Noticeable upward arm movements are not as present in many other sports, hence the low intensity of the activation in the rest of the map. Conversely, the ``soccer kick'' basis pose is mostly dominant in the top-left area and the heatmap is diffused, consistent with the observation that most poses contain some movement of the legs.

%%%%%%%%%%%%%%%%%%%%%%%%%%%%%%%%%%%%%%%%%%%%%%%%%%%%%%%%%%%%%%%%%%%%%%%%%%%%%%%%%%%%%%%%%%
%%%%%%%%%%%%%%%%%%%%%%%%%%%%%%%%%%%%%%%%%%%%%%%%%%%%%%%%%%%%%%%%%%%%%%%%%%%%%%%%%%%%%%%%%%
%% Conclusions
%%%%%%%%%%%%%%%%%%%%%%%%%%%%%%%%%%%%%%%%%%%%%%%%%%%%%%%%%%%%%%%%%%%%%%%%%%%%%%%%%%%%%%%%%%
%%%%%%%%%%%%%%%%%%%%%%%%%%%%%%%%%%%%%%%%%%%%%%%%%%%%%%%%%%%%%%%%%%%%%%%%%%%%%%%%%%%%%%%%%%

\section{Conclusion and Future Directions}
In this paper, we proposed a model for learning the primitive movements underlying human actions (movemes) from a set of static 2-D poses obtained from images taken at various angles of view. The bases poses are rotation-invariant and learned through a modified latent matrix factorization that intrinsically accounts for geometric properties inherent to viewing angle variability. The approach can be trained efficiently, requires modest effort to identify a reasonable initialization, and yields very good generalization on unseen data.

We investigated the practical use of the learned representation for applications such as activity recognition and inference of action dynamics, observing significantly better performance compared to conventional baselines that do not account for variability of viewing angles. We used the bases poses for synthetic generation of movements, and explored how specific poses are mapped to different parts of the high-dimensional manifold of human motion.

One desirable property of our algorithm is that it is complementary to existing latent factor, pose estimation and feature extraction approaches, and may be used in combination with them to yield a better overall rotation-invariant representation.

An interesting future direction of investigation would be to use the proposed model in a semi-supervised setting where there is some availability of true three-dimensional data along with a large collection of two-dimensional joint locations.

Other possible extensions of our work are: learning to morph actions and synthesize \textit{unseen} actions from the set of extracted movemes; inferring the location of occluded or missing joints based on the position of the visible ones; applying these techniques to large-scale datasets \cite{lin2014microsoft} in conjunction with fine grained annotations of the performed actions \cite{BMVC2015_52, krishnavisualgenome} to gain new insights on the structure, complexity, and duration of human behaviour.

%%%%%%%%%%%%%%%%%%%%%%%%%%%%%%%%%%%%%%%%%%%%%%%%%%%%%%%%%%%%%%%%%%%%%%%%%%%%%%%%%%%%%%%%%%
%%%%%%%%%%%%%%%%%%%%%%%%%%%%%%%%%%%%%%%%%%%%%%%%%%%%%%%%%%%%%%%%%%%%%%%%%%%%%%%%%%%%%%%%%%
%% Bibliography
%%%%%%%%%%%%%%%%%%%%%%%%%%%%%%%%%%%%%%%%%%%%%%%%%%%%%%%%%%%%%%%%%%%%%%%%%%%%%%%%%%%%%%%%%%
%%%%%%%%%%%%%%%%%%%%%%%%%%%%%%%%%%%%%%%%%%%%%%%%%%%%%%%%%%%%%%%%%%%%%%%%%%%%%%%%%%%%%%%%%%

\bibliographystyle{IEEEtran}
\bibliography{IEEEabrv,egbib}

\end{document}